\documentclass[conference]{IEEEtran}
\IEEEoverridecommandlockouts
\usepackage{cite}
\usepackage{amsmath,amssymb,amsfonts}
\usepackage{graphicx}
\usepackage{textcomp}
\usepackage{xcolor}
\usepackage[font=small]{caption}
\usepackage{algorithm}
\usepackage{algpseudocode}
\usepackage{siunitx}

\def\BibTeX{{\rm B\kern-.05em{\sc i\kern-.025em b}\kern-.08em
    T\kern-.1667em\lower.7ex\hbox{E}\kern-.125emX}}
\begin{document}

\title{Deep Learning for Modeling and Dispatching  Hybrid Wind Farm Power Generation\\

\thanks{This work was funded in part by NSF REU Award 2243980}
}

\author{\IEEEauthorblockN{Zach Lawrence}
\IEEEauthorblockA{\textit{Computer Science} \\
\textit{University of California, San Diego}\\
La Jolla, CA, United States \\
zclawrence@ucsd.edu}
\and
\IEEEauthorblockN{Jessica Yao}
\IEEEauthorblockA{\textit{Statistics and Data Science} \\
\textit{University of California, Los Angeles}\\
Los Angeles, CA, United States \\
jessicayao26@ucla.edu}
\and
\IEEEauthorblockN{Chris Qin}
\IEEEauthorblockA{\textit{Mechanical Engineering} \\
\textit{Washington State University, Vancouver}\\
Vancouver, WA, United States \\
chris.qin@wsu.edu}
}

\maketitle


\begin{abstract}
Wind farms with integrated energy storage, or hybrid wind farms, are able to store energy and dispatch it to the grid following an operational strategy. For individual wind farms with integrated energy storage capacity, data-driven dispatch strategies using localized grid demand and market conditions as input parameters stand to maximize wind energy value. Synthetic power generation data modeled on atmospheric conditions provide another avenue for improving the robustness of data-driven dispatch strategies. To these ends, the present work develops two deep learning frameworks: COVE-NN, an LSTM-based dispatch strategy tailored to individual wind farms, which reduced annual COVE by 32.3\% over 43 years of simulated operations in a case study at the Pyron site; and a power generation modeling framework that reduced RMSE by 9.5\% and improved power curve similarity by 18.9\% when validated on the Palouse wind farm. Together, these models pave the way for more robust, data-driven dispatch strategies and potential extensions to other renewable energy systems. 
\end{abstract}


\section*{Abbreviations}
\addcontentsline{toc}{section}{Abbreviations}
\begin{IEEEdescription}[\IEEEusemathlabelsep\IEEEsetlabelwidth{$V_1,V_2,V_3$}]
\item[$AEP$] Annual energy production.
\item[$BOS$] Balance of system.
\item[$CAES$] Compressed air energy storage.
\item[$CAPEX$] Capital expenditure.
\item[$CNN$] Convolutional neural network.
\item[$COVE$] Cost of valued energy.
\item[$CRPS$] Continuous ranked probability score.
\item[$FCR$] Fixed charge rate.
\item[$LCOE$] Levelized cost of energy.
\item[$LSTM$] Long short-term memory (module).
\item[$NQF$] Neural quantile function (e.g., in NQF-RNN).
\item[$OPEX$] Operational expenditure.
\item[$RNN$] Recurrent neural network.
\item[$RMSE$] Root mean squared error.
\item[$RTE$] Round-trip efficiency.
\item[$VF$] Value factor.
\item[$_{S}$] Subscript referring to storage component (e.g., $CAPEX_{S}$).
\item[$_{WF}$] Subscript referring to wind farm component (e.g., $CAPEX_{WF}$).
\end{IEEEdescription}

\section{Introduction}
Wind power has seen a marked increase in cost-efficiency and growth in the last 15 years, playing a key role in electricity generation worldwide \cite{renewables}. To meet growing energy demands, development has accelerated on large-scale wind farms consisting of clusters of wind turbines both onshore and offshore. As regions with favorable atmospheric conditions for wind energy production see increasing wind farm development, engineers are beginning to focus on addressing intermittency, enhancing power grid integration, aligning generation with user demand, and ultimately improving the overall profitability of wind farms \cite{COVE}. \textcolor{black}{Furthermore, decarbonization efforts increase pressure for future high-penetration renewable energy sources to respond to energy demands with stable, low-intermittency power dispatch \cite{high-penetration}.} With these goals in mind, wind farms with integrated energy storage systems, known as hybrid wind farms, are being explored. 

With hybrid wind farms, surplus wind energy may be stored and later regenerated to the power grid following a dispatch strategy, the planned scheduling of power release to match grid demand and market conditions, with the aim of maximizing wind energy value. \textcolor{black}{Dispatch strategies optimized around the storage capabilities of hybrid wind farms provide a potential avenue for stable, low-intermittency power dispatch.} Because the price of electricity across the power grid fluctuates rapidly, slight variations to dispatch strategies can accumulate to significant differences in annual energy value. As such, data-driven dispatch strategies often use a wealth of time-dependent covariates from electrical market data, atmospheric data, and operational data sourced from wind farms directly. Due to lacking data availability in these areas, modeling and forecasting of atmospheric data and wind power generation are long-standing areas of interest, as they can provide synthetic time series for dispatch strategies and operational analysis. 

In power generation modeling, the aerodynamic interaction among wind turbines leads to velocity deficits, causing turbulence and wake effects that reduce the overall efficiency of wind farms. Wake losses arise because turbines extract momentum from incoming wind, creating regions of reduced wind speed that propagate downstream and impact turbines located in the flow direction. Capturing the combined effects of atmospheric variability and unsteady wake dynamics has proven challenging for traditional closed-form analytical approaches. As an alternative or complement to traditional models, many recent approaches incorporate empirical or data-driven strategies to represent turbulence and wake interactions in power generation modeling. When integrated with dispatch strategies, existing power generation models often struggle to fully exploit temporal dependencies in atmospheric data, limiting their ability to provide smooth and accurate forecasts over the time horizons required for grid scheduling. 

Dispatch strategies are often developed on a case-by-case basis, in which some auxiliary metric is optimized alongside energy cost, often towards applications across an entire grid consisting of many power systems \cite{yu}, \cite{thukaram}, \cite{peterson}. This choice in design limits the ability of many dispatch strategy methodologies to generalize across renewable energy sources and infrastructure contexts. Additionally, because many dispatch strategies require grid-level knowledge of multiple power systems, their applicability is less than ideal for small scale experimentation and validation on individual wind farms. The objective of this work is to develop data-driven methodologies that directly address two central challenges in wind energy research: 
\begin{itemize}
    \item[(i)] Designing power dispatch strategies that enhance the value of wind power to the grid for individual wind farms\textcolor{black}{, with a focus on stable power dispatch and reduced grid intermittency, anticipating high-penetration renewable energy needs}.
    \item[(ii)] Improving wind power generation modeling \textcolor{black}{by leveraging temporal dependencies in wind speed and power generation.}
\end{itemize}
By focusing on these objectives, this work aims to provide practical tools for reducing grid intermittency and improving the economic performance of wind farms.

To the best of the authors’ knowledge, this is the first work to propose a deep-learning-based dispatch strategy tailored to an individual \textcolor{black}{wind power system}. Specifically, the study introduces COVE-NN, a recurrent neural network (RNN) that uses long-short-term memory (LSTM) modules for power dispatch, and demonstrates its effectiveness through a case study of the Pyron wind farm in Texas, achieving a 32.3\% reduction in the annual cost of valued energy (COVE) compared to a conventional baseload strategy over 43 years of simulated data. In addition, this work develops a deep learning framework for wind power generation modeling that achieved a 9.5\% reduction in root mean squared error (RMSE), a 1.9\% improvement in cross correlation, and an 18.9\% improvement in power curve similarity over baseline models evaluated on historical data from the Palouse wind farm. Together, these contributions establish a novel methodological foundation for integrating generation modeling with dispatch strategies in future research and applications.

\section{Related Work}
There are a variety of approaches in the literature for power generation modeling based on probabilistic, physics-informed, and classical machine learning methodologies. Karp and Qin propose a data-driven probabilistic methodology for modeling wind-farm specific power generation given wind speed \cite{ivan}. Soroudi, Aien, and Ehsan propose a probabilistic methodology for modeling grid-wide wind power generation \cite{soroudi}. Sweetman and Dai develop a data-driven methodology for transforming ideal zero-turbulence power curves to practical, turbulence-modeling power curves using random process theory \cite{sweetman}. Bardal and Saetran experimentally validate physics-informed, deterministic power curve models with varying treatments of turbulence \cite{bardal}. Clifton and Wagner develop a data-driven random forest model that accounts for turbulence in wind power curves, in which their approach outperforms deterministic, physics-informed turbulence models \cite{clifton}. 

Power dispatch strategies in the literature predominantly make use of classical optimization and computational stochastic techniques to improve the efficiency of conventional peaking and baseload strategies, often applied across a grid containing multiple power systems. Yu, Wu, and Bhattacharya use particle swarm optimization to minimize microgrid dispatch cost and emission pollution \cite{yu}. Thukaram and Parthasarathy develop an algorithm for optimal voltage stability and cost minimization given control over multiple power systems and grid parameters \cite{thukaram}. Peterson, Bendtsen, and Stoustrup develop a methodology for optimizing dispatch in a grid with multiple renewable energy sources that ensures stability under future load imbalances using quadratic programming \cite{peterson}. For a more fine-grained analysis of dispatch strategies for individual power systems, Zurita, Mata-Torres, Cardemil, Guédez, Escobar analyze the cost efficiency of multiple dispatch strategies and integrated storage solutions for solar power systems, finding that the baseload dispatch strategy incurred the lowest energy costs \cite{zurita}. \textcolor{black}{Previous studies have explored deep-learning-based dispatch optimization for hybrid systems using LSTM-based RNN architectures \cite{2stage-dispatch}, \cite{dl-microgrid}, \cite{wind-hydrogen-dispatch}. However, these approaches primarily focus on multi-system or grid-level optimization, while the present work uniquely targets an individual wind power system with an unsupervised COVE-based loss formulation.}

For the purposes of developing data-driven power generation and dispatching models and strategies, the aforementioned works suffer from pitfalls that reduce their practical applicability for real-time deployments. For power generation modeling, many approaches in the literature do not take advantage of temporal dependencies in atmospheric data such as wind speed. For power dispatch, much of the literature is designed around the control of multiple power systems, and is of limited applicability in real-time, single wind farm settings. In pursuing data-driven models that can address these issues, work carried out in wind energy and wind speed forecasting frequently makes use of deep learning architectures that promote the latent use of temporal dependencies in model predictions, as well as real-time applicability in the form of fast inference times with robustness to noisy or incomplete data. Wang \textit{et al}. use convolutional neural networks (CNN) for probabilistic wind energy forecasting based on spatial information \cite{wang}. More recently, recurrent neural networks (RNN) have been used for forecasting either as a standalone architecture or as a supplement to CNN-based approaches, as explored in work by Afrasiabi, Mohammadi, and Rastegar \cite{afrasiabi} and in reviews of contemporary deep learning methods for renewable energy forecasting carried out by Wang, Lei, Zhang, Zhou, and Peng \cite{wang_review}.
 
\section{Methodology}
\subsection{Deep Learning Methodology}
For both the power generation and dispatch models, a common network architecture was devised, which was modified under both cases to fit problem constraints. This common network is an LSTM-based recurrent neural network in which the activations exiting the LSTM are forwarded to a feedforward network that terminates in a single neuron \cite{LSTM}. The activation of this single neuron represents the output for both problem cases $y_t$, and, given the input time series to the network has not terminated, $y_t$ is forwarded to the LSTM as an autoregressive input alongside the next set of multivariate inputs in the time series. For the first set of multivariate inputs in a time series, the autoregressive input is predetermined per problem case. In both problem cases, the input time series contains multiple covariates $x_{0,t}, x_{1,t}, \dots, x_{n,t}$ at a given timestep $t$ in the series. 

\begin{figure}[h!]
    \centering
    \includegraphics[width=1\linewidth]{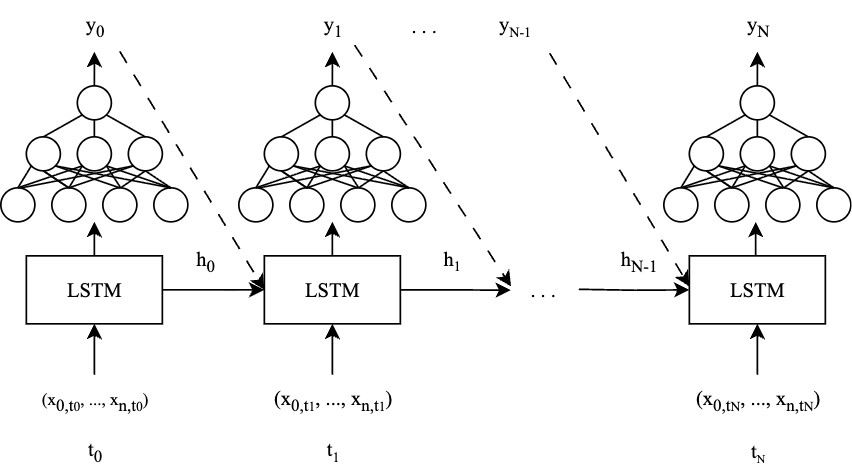}
    \caption{Baseline Network for Adaptation to Problem Constraints}
    \label{fig:common-net}
\end{figure}
Between recurrent passes through the network, the LSTM forwards hidden state and cell state tensors, contained in $h_t$, which allow the network to learn temporal dependencies from the input time series. 

\subsubsection{Power Generation Modeling}
In the power generation modeling case, the common neural network is extended into the Neural Quantile Function with Recurrent Neural Networks (NQF-RNN) framework \textcolor{black}{for probabilistic wind power forecasting \cite{NQFRNN}. This work represents the first application of the NQF-RNN to wind power generation modeling, in which the framework is adapted to capture the stochastic characteristics of wind energy data.} The model parametrizes a monotonic neural quantile function that maps quantile functions to predictions, allowing the network to learn and predict a probability distribution at each timestep. Training is performed using a continuous ranked probability score (CRPS)-based loss function with a bias term that penalizes deviations between predicted and observed means. During training, a set of quantile values are selected, and the model makes a prediction for each quantile on each training sample. These predictions approximate the model's predicted probability distribution conditioned on the input covariates, and this distribution is passed to the CRPS loss function for learning. 

The output from the feedforward network, which acts as the neural quantile function, is $p_t$ (MW), the modeled power generation at timestep $t$. The inputs to the model are the wind speed $v_t$ (m/s) at timestep $t$, along with the previous timestep prediction $p_{t-1}$, which is passed to the model autoregressively. In addition, the neural quantile function receives an input quantile level, $\alpha(t) \in (0,1)$, which specifies the quantile from which the model predicts $p_t$, thereby modeling quantile sampling from a probability distribution of $p_t$ conditioned on $v_t, p_{t-1}$.

After model training is complete, a Brownian smoothing approach is used during inference to sample values from the predictive distributions to promote a smooth time series. \textcolor{black}{This is an additional procedure introduced in this work to reduce excessive noise in the generated power values to better reflect temporal continuity.} This approach consists of generating a sequence of quantile levels through a random walk of the standard normal distribution that progresses smoothly over time, reflecting at the boundaries when necessary to ensure $\alpha(t) \in (0, 1)$ to avoid invalid quantiles. This procedure helps to ensure that successive samples vary smoothly while remaining probabilistically consistent, reflecting the inherent temporal continuity of wind speeds and power generation. The smoothing equation is defined as:
\begin{equation}\label{eq:brown}
\alpha(t) = \alpha(t-1) + \lambda N(0,1) + \gamma(0.5 - \alpha(t))
\end{equation}

\begin{itemize}
  \item[] $\alpha(t)$: quantile level input at timestep $t$
  \item[] $\lambda$: smoothness factor
  \item[] $\gamma$: drift factor pulling progress back to median at 0.5
\end{itemize}
\begin{figure}[h!]
    \centering
    \includegraphics[width=1\linewidth]{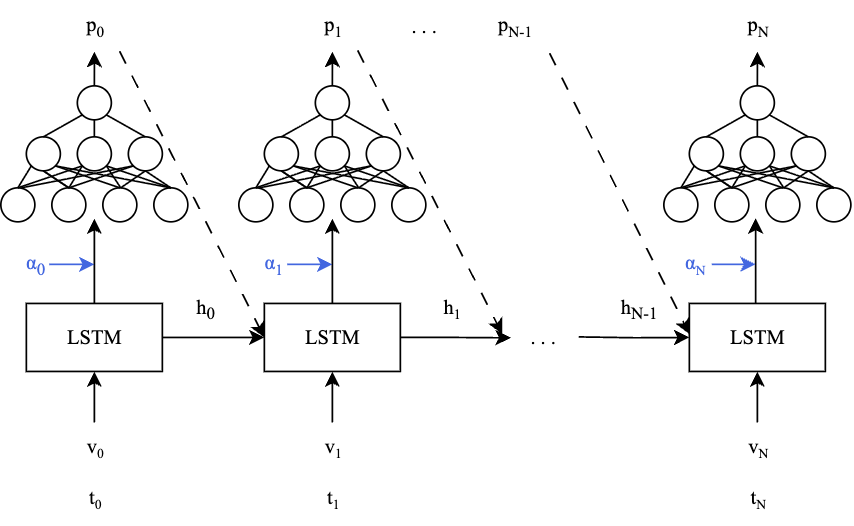}
    \caption{NQF-RNN Model Architecture. The modification to the common architecture is highlighted in blue, whereby an additional quantile level input $\alpha$ is passed to the feedforward network, known as the neural quantile function (NQF).}
    \label{fig:nqf-rnn}
\end{figure}

WE-Validate, a framework for wind power validation \cite{wevalidate}, is used to evaluate model performance. From this framework, two standard metrics were used: root mean squared error (RMSE) and cross correlation. In addition, \textcolor{black}{this study introduces a} third metric called power curve similarity, designed to capture the distributional patterns of power generation not reflected in point-wise errors or correlations.

First, root mean squared error (RMSE) measures the average magnitude of errors between predictions and observed values, where a lower score indicates better fit.
\begin{equation}\label{eq:rmse}
    RMSE = \sqrt{\frac{\sum^{n}_{i=1}(\hat{y}_i - y_i)^2}{n}}
\end{equation}

Second, cross correlation (also called the Pearson Correlation Coefficient) measures the relation of two time series, where a higher score indicates better fit. 
\begin{equation}\label{eq:cross-cor}
    r = \frac{\sum^{N}_{i=1} (x_i-\bar{x})(y_i-\bar{y})}{\sqrt{\sum^{N}_{i=1}(x_i-\bar{x})^2 \sum^{N}_{i=1}(y_i-\bar{y})^2 }}
\end{equation}

Finally, the power curve similarity metric compares the shape of the joint distribution of wind speed $v$ and power $p$ between predicted and historical data. The joint distributions of wind speed and power were estimated, $f_{hist}(v, p)$ for historical data and $f_{pred}(v, p)$ for predicted data. The similarity is quantified using Jensen-Shannon Divergence (JSD), with values closer to $1$ indicating better performance:
\begin{equation}\label{eq:jsd}
Similarity = 1 - JSD(f_{hist}(v, p) || f_{pred}(v, p))
\end{equation}

\subsubsection{Power Dispatch Strategy}
For the power dispatch strategy, the aforementioned common network is modified in its autoregressive behavior to model integrated storage constraints. Integrated storage in a hybrid wind farm is constrained to the following parameters: \textit{rating}, $R_S$, the maximum power that can be charged to or discharged from storage at a given time step, and \textit{duration}, the number of timesteps in which, with no stored energy, the system can operate at its full charge rating until reaching its storage capacity. Note that storage \textit{capacity}, $C_S$, is equivalent to \textit{rating} $*$ \textit{duration}. Based on these parameters, additional dependent parameters can be determined per storage system, including round-trip efficiency ($RTE_S$), capital expenditures ($CAPEX_S$), and operational expenditures ($OPEX_S$). The covariates inputted to the model $x_{0,t}, \dots, x_{n,t}$ are as follows: 

\begin{itemize}
  \item[] $g_t$: wind power generated (MW) at timestep $t$
  \item[] $p_t$: electricity price (LMP, \$/MWh) at timestep $t$
  \item[] $s_t$: total stored energy (MWh) at timestep $t$
  \item[] $u_t$: user electrical load (MW) at timestep $t$
\end{itemize}

For a given timestep $t$, $r_t$ is predicted by the model, representing the amount of power to dispatch to the grid at timestep $t$. To reduce model complexity, $r_t$ is treated as a \textit{capacity factor} (CF) scaled by the rated capacity of the wind farm of interest $C_{WF}$. That is, $0 \leq r_t \leq 1$, where the power dispatched to the grid is $r_t*C_{WF}$. In order to ensure dispatched power and stored energy respects the integrated storage constraints, Algorithm \ref{alg:post} determines the final power to be dispatched to the grid at timestep $t$, $r'_t$ and the total stored power going into the next timestep, $s_{t+1}$ after model prediction of $r_t$.
\begin{algorithm}[h!]
\caption{Power Dispatch Model Post-Processing}\label{alg:post}
\begin{algorithmic}[1]
\Require $0 \leq r_t \leq 1$
\Require $0 \leq s_t \leq C_S$
\State $r'_t \gets $min$(r_t*C_{WF}, g_t+s_t)$
\State $r_{regen}\gets $min$($max$(r'_t - g_t,0), R_S)$
\State $r_{direct}\gets $max$(r'_t-r_{regen}, 0)$
\State $r'_t\gets r_{direct}+(r_{regen}*RTE_S)$
\State $s_{t+1}\gets $max$($min$(s_t+$min$(g_t-r'_t,R_S), C_S),0)$
\end{algorithmic}
\end{algorithm}

\begin{figure}[h!]
    \centering
    \includegraphics[width=1\linewidth]{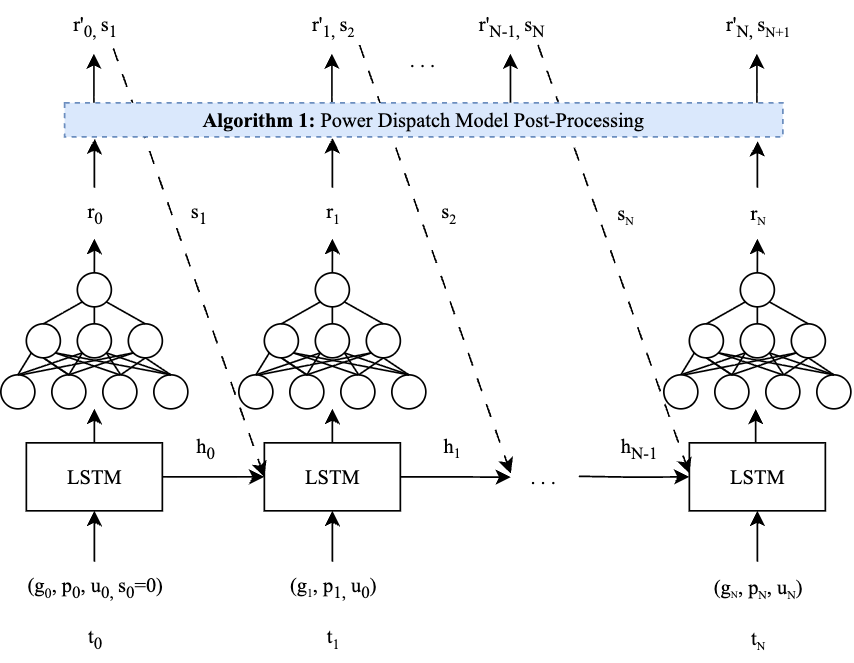}
    \caption{COVE-NN Model Architecture. The modification from the common architecture is highlighted in blue, whereby outputs from the network are passed through Algorithm \ref{alg:post}. The resulting $r_t'$ values are recorded and the $s_{t+1}$ values are forwarded recurrently to the  LSTM alongside the $t+1^{th}$ set of input covariates.}x
    \label{fig:cove-nn}
\end{figure}
The goal of COVE-NN is to maximize the annual value of wind energy, or, equivalently, minimize the annual cost of wind energy. Standard metrics for evaluating the cost of energy, such as levelized cost of energy (LCOE), do not take into account the market-valued price of energy.

\begin{equation}
  \label{eq:lcoe}
  \begin{gathered}[b]
     LCOE(r',R_S,D_S)=\frac{[CAPEX*FCR + OPEX]_{WF}}{\sum_ir'_i}
    \cr + \frac{[CAPEX(R_S,D_S)*FCR + OPEX(R_S,D_S)]_S}{\sum_ir'_i}
  \end{gathered}
\end{equation}

This simplification poses problems for precise cost comparisons across dispatch strategies, especially when conventional strategies, such as peaking dispatch strategies, use market-valued electrical prices as a key parameter to power dispatch decisions. Therefore, this work opts for an alternative metric, cost of valued energy (COVE), which accounts for the market-valued price of electricity \cite{COVE}.

\begin{equation}
  \label{eq:cove}
  \begin{gathered}[b]
     COVE(r', p, R_S, D_S)=\frac{[CAPEX*FCR + OPEX]_{WF}}{\sum_ir'_i*p_i} 
    \cr + \frac{[CAPEX(R_S,D_S)*FCR + OPEX(R_S,D_S)]_S}{\sum_ir'_i*p_i}
  \end{gathered}
\end{equation}

In Equations \ref{eq:lcoe} and \ref{eq:cove}, $CAPEX_{WF}$ contains the balance of system costs ($BOS$), and $CAPEX_S$, $OPEX_S$ are parameterized by storage rating and duration $R_S$, $D_S$, respectively. In order to reduce reliance on conventional power dispatch strategies and thereby promote the emergence of new strategies, COVE-NN is trained using unsupervised learning. The loss function, defined in Equation \ref{eq:unsup-loss}, contains the COVE term to be minimized, as well as two additional tunable, adaptive terms.
\begin{equation}
\label{eq:unsup-loss}
    \begin{gathered}[b]
         \mathcal{L}(r',p,s,g)=COVE(r',p,R_S,D_S)\\
         +\frac{\lambda}{max(t-t_a,1)^\Lambda}((\gamma (s*\frac{p}{\bar{p}}))^\Gamma + (\omega *max(r'-\bar{g},0)\frac{\bar{p}}{p})^\Omega)
      \end{gathered}
\end{equation}

During initial experiments, it was found that exclusively training the model on a COVE-based loss resulted in early convergence, in which COVE-NN massively under-utilized storage. Taking inspiration from Bayati and Jabbarvaziri, the two adaptive loss terms penalize COVE-NN for deviating from \textit{baseload} and \textit{peaking} dispatch strategies, which prevent early convergence to spurious local minima by introducing more convex terms to the loss function \cite{bayati}. In discussions of baseload, this work refers to a peak-shaving implementation of a deterministic dispatch algorithm \cite{reina}. In this strategy, power is dispatched at a fixed average level to maintain a consistent and stable supply. While originally rooted in non-renewable generation, baseload-style dispatch remains a useful heuristic for low-intermittency dispatch systems systems and serves as a reference point for evaluating more dynamic strategies \cite{reina}, \cite{fasihi}, \cite{iori}. In comparison, peaking dispatch strategies dynamically release bursts of energy with the intent of maximizing energy value, intermittently releasing power when user demand and market price are exceedingly high.

The adaptive loss terms promote the emergence of data-driven strategies, latent "hybrids" of conventional dispatch strategies learned by the model. \textcolor{black}{In extending conventional baseload and peaking dispatch strategies, COVE-NN was able to achieve stable power dispatch in alignment with the dual goals of maximizing energy value and reducing grid intermittency.} By adaptively mitigating the aforementioned baseload and peaking penalties in higher training epochs, COVE-NN is able to finetune the hybrid dispatch strategy it has learned in early epochs, further minimizing COVE. Table \ref{tab:loss_hp} defines the tunable hyperparameters used in Equation \ref{eq:unsup-loss}. Note that $\bar{g},\bar{p}$ refer to the mean power generation and electricity price, respectively, over the input vectors $g,p$ in Equation \ref{eq:unsup-loss}.

\begin{table}[h!]
    \centering
    \begin{tabular}{|c|c|}
    \hline
        \textbf{Hyperparameter} &  \textbf{Description}\\
        \hline
        $\gamma$ & Peaking penalty pre-factor \\
        \hline
        $\Gamma$ & Peaking penalty degree\\
        \hline
        $\omega$ & Baseload penalty pre-factor\\
        \hline
        $\Omega$ & Baseload penalty degree\\
        \hline
        $ \lambda$ & Time-adaptive pre-factor\\
        \hline
        $ \Lambda$ & Time-adaptive degree\\
        \hline
        $ t_a$ & Epochs before time-adaptive scaling applied \\
        \hline
    \end{tabular}
    \caption{COVE-NN Unsupervised Loss Hyperparameters}
    \label{tab:loss_hp}
\end{table}

For model evaluation following training, some additional metrics were captured alongside annual COVE to provide a comprehensive overview of COVE-NN performance. These metrics include annual energy production (AEP), annual power curtailment, average storage utilization relative to capacity, and annual value factor (VF). 

\begin{equation}
    VF=\frac{\sum_{i}^Ng_ip_i}{[\sum_i^Ng_i]*[\frac{1}{N}\sum_i^Np_i]}
\end{equation}

In particular, value factor is a useful metric for determining a dispatch strategy's tendency to dispatch power at a constant rate, a desirable behavior for reducing intermittency in individual power systems.

\subsection{Case Study: Pyron Wind Farm}
In order to verify the above methodology, the Pyron wind farm located in Roscoe, Texas, was selected for a case study due to the high data availability provided by the Electric Reliability Council of Texas (ERCOT) as well as the high wind penetration in the area. The Pyron wind farm is rated at 249 MW with 166 GE 1.5 MW turbines \cite{GridInfo}. The Herbie Python package \cite{Herbie} was used to download hourly nodal wind speeds from High-Resolution Rapid Refresh (HRRR), spanning from July 2014 to December 2023. Simulated hourly wind power generation profiles for the Pyron site, nodal locational marginal prices (LMP), and zonal user electrical loads were obtained from ERCOT \cite{ERCOT}.

\begin{table}[h!]
    \centering
    \begin{tabular}{|c|c|c|c|}
    \hline
        \textbf{Dataset} & \textbf{Time Span} & \textbf{Source} & \textbf{Coverage}\\
        \hline
        Wind Speed & 2014 - 2023 & HRRR \cite{Herbie} & Nodal\\
        \hline
        Wind Power Generation & 1980 - 2023 & ERCOT \cite{ERCOT} & Nodal\\
        \hline
        Electricity Price (LMP) & 2010 - 2023 & ERCOT \cite{ERCOT} & Nodal\\
        \hline
        User Electrical Load & 2018 - 2023 & ERCOT \cite{ERCOT} & Zonal\\
        \hline
    \end{tabular}
    \caption{Pyron Case Study: Dataset Specifications. The wind power generation was modeled by ERCOT, and is not historical data. The electricity price, user electrical load, and wind speed data are all historical.}
    \label{tab:my_label}
\end{table}

In particular, the training and evaluation dataset for the power generation model consisted of all available wind speeds (2014 - 2023) and simulated wind power generation spanning the same time frame (2014 - 2023). The training and evaluation dataset for COVE-NN consisted of all available power generation (1980 - 2023), with electricity price and user load data repeated in sequence to fit all 43 years of data. For both datasets, a train-test split was carried out whereby 70\% of each dataset was used for training, and the other 30\% was used for testing and validation.

\textcolor{black}{Although this case study focuses on the Pyron Wind Farm, similar validation could be conducted for other sites. Obtaining high-resolution, hourly datasets that simultaneously include a single wind farm's generation data and its associated nodal locational marginal price (LMP) records is difficult, as such combined datasets are almost nonexistent in the public domain. Fortunately, the Electric Reliability Council of Texas (ERCOT) provides modeled wind-farm generation data based on HRRR atmospheric inputs, together with over a decade of continuous nodal LMP records. The Pyron Wind Farm was selected not only because of this data availability but also because its rated capacity of approximately 250 MW is representative of modern large-scale wind farm design. Moreover, Pyron is planned to incorporate energy storage capacity from RWE's Texas Wave Project ($\approx$ 10 MW / 50 MWh) and the under-construction Texas Wave II project ($\approx$ 30 MW / 30 MWh), which will provide co-located battery systems directly coupled to the Pyron site \cite{rwe}. Once operational and publicly documented, these assets will offer valuable data for future validation and analysis. In addition, ongoing initiatives such as NREL's AWAKEN experiment and national resources like NWPDB and EIA-930 (although regional rather than nodal) are expected to enable further case studies as higher-resolution datasets become available \cite{awaken}.}

\subsubsection{Power Generation Model Evaluation}

To evaluate the ability of the power generation model to learn from smaller atmospheric datasets and establish baseline models for comparison, a second validation dataset was used from the Palouse Wind Farm using the WE-Validate framework \cite{wevalidate}. This framework provides historical power data and modeled results from two \textcolor{black}{established} baseline models, \textcolor{black}{NWPDB and PLUSWIND,} allowing for direct comparison against the model developed in this work. Additionally, to generate model predictions on this dataset, wind speed data was independently obtained from HRRR over the region spanned by the Palouse wind farm. 

\begin{table}[h]
    \centering
    \begin{tabular}{|c|c|c|}
    \hline
        \textbf{Dataset} & \textbf{Time Span} & \textbf{Source} \\
        \hline
        Historical Wind Speed & Jan-Dec 2018 & HRRR \cite{Herbie}\\
        \hline
        Historical Wind Power Generation & Jul-Dec 2018 & EIA-930 \cite{wevalidate} \\
        \hline
        NWPDB Model & Jul-Dec 2018 & NWPDB  \cite{wevalidate} \\
        \hline
        PLUSWIND Model & Jan-Dec 2018 & PLUSWIND  \cite{wevalidate}\\
        \hline
    \end{tabular}
    \caption{Palouse Wind Farm: Dataset Specifications}
    \label{tab:palouse-dataset}
\end{table}

The power generation model was retrained and evaluated using a dataset consisting of historical wind speed and power generation from July to December of 2018. \textcolor{black}{PLUSWIND provides hourly power generation estimates using meteorological input from the High-Resolution Rapid Refresh (HRRR) model. It uses turbine-specific power curves adjusted for air density to produce deterministic power generation estimates \cite{wevalidate}. NWPDB (National Wind Power Database) is a national synthetic database that also uses HRRR meteorological data and plant configurations from the EIA-860 and U.S. Wind Turbine Database to generate} a probabilistic model with uncertainty quantification for power generation \cite{wevalidate}.

\subsubsection{Power Dispatch Strategy Evaluation}
To evaluate the performance of COVE-NN, \textit{baseload} was selected as a baseline dispatch strategy for comparison, due to its historical prevalence in \textcolor{black}{stable, low-intermittency, low-cost non-renewable energy sources} \cite{fasihi}, \cite{baseload-renewables}. \textcolor{black}{Baseload was chosen as opposed to more dynamic peaking-mode dispatch strategies because, although baseload is less common in current renewable energy infrastructure, it provides a useful benchmark against which to evaluate low-intermittency, stable power sources in anticipation of high-penetration renewable infrastructure needs, which this work aims to explore.} Because COVE-NN requires a choice of integrated energy storage hyperparameters, a search was conducted over storage systems and their parameters to determine which storage system optimized baseload performance. The storage systems and parameters described in Table \ref{tab:storage-params} were sourced from the Pacific Northwest National Laboratory (PNNL) \cite{pnnl}. Alongside the values provided in Table \ref{tab:storage-params}, PNNL also provided round-trip efficiencies ($RTE_S$), capital expenditures ($CAPEX_S$), and operational expenditures ($OPEX_S$) parameterized by rating and duration per storage system, which were used in Algorithm \ref{alg:post} and COVE calculations. Additionally, for all COVE calculations, $CAPEX_{WF}$, $OPEX_{WF}$, and $FCR$ were provided by the National Renewable Energy Laboratory (NREL) \cite{nrel}.
\begin{table}[h!]
    \centering
    \begin{tabular}{|c|c|c|}
    \hline
    \textbf{Storage} & \textbf{Ratings (MW)} & \textbf{Durations (hrs)}\\
    \hline
    Lithium-Ion & \{100, 1000\} & \{2, 4, 6, 8, 10, 24, 100\}\\
       \hline
       Hydropower & \{100, 1000\} & \{4, 10, 24, 100\}\\
       \hline
       CAES & \{100, 1000\} & \{4, 10, 24, 100\}\\
       \hline
       Hydrogen & \{100, 1000\} & \{10, 24, 100\}\\
       \hline
       Gravitational & \{100, 1000\} & \{2, 4, 6, 8, 10, 24, 100\}\\
       \hline
       Thermal & \{100, 1000\} & \{4, 6, 8, 10, 24, 100\}\\
       \hline
    \end{tabular}
    \caption{Storage Parameters Searched across Systems}
    \label{tab:storage-params}
\end{table}
\begin{table}[h!]
    \centering
    \begin{tabular}{|c|c|}
    \hline
        \textbf{Hyperparameter} &  \textbf{Search Range}\\
        \hline
        $\gamma$  & $(0,3)$\\
        \hline
        $\Gamma$ & $(0,3)$\\
        \hline
        $\omega$  & $(0,5)$\\
        \hline
        $\Omega$ & $(0,5)$\\
        \hline
        $ \lambda^*$ & $1$\\
        \hline
        $ \Lambda$ & $(\frac{1}{8}, \frac{1}{4})$\\
        \hline
        $ t_a^*$ & $8$\\
        \hline
    \end{tabular}
    \caption{COVE-NN Hyperparameter Search Ranges. Note that $\lambda,t_a$ were not searched over in order to mitigate unintended changes to loss penalties during early epochs, and are included here for clarity.}
    \label{tab:hp_search}
\end{table}

After completing the integrated storage hyperparameter search using baseload, a hyperparameter search was conducted on COVE-NN using the storage system that minimized baseload annual COVE. The hyperparameter search on COVE-NN randomly initialized COVE-NN hyperparameters from the ranges shown in Table \ref{tab:hp_search}. After 10 epochs, the average annual COVE computed from in-training COVE-NN predictions on the validation dataset was compared to the minimum average annual COVE computed during the hyperparameter search. If the in-training COVE-NN minimized the average annual COVE over the validation dataset, training continued and the minimum COVE was updated accordingly. Otherwise, training was terminated early and a new set of hyperparameters was gathered for training a the next model in the search. 

\section{Results}

\subsection{Power Generation Modeling}
The performance of the NQF-RNN model trained on different time series lengths (hours) per training sample was evaluated to determine the effect of temporal context on predictive accuracy. Table \ref{nqfrnn-comparison} contains the results across all four time series lengths tested, using the RMSE, cross correlation, and power curve similarity metrics. Among all models tested, the 168-hour sequence model achieved the best performance across all metrics, with an RMSE of 0.189, cross correlation of 0.799, and a power curve similarity of 0.803. 

\begin{table}[h!]
    \centering
    \begin{tabular}{|c|c|c|c|}
    \hline
        \textbf{Hours} & \textbf{RMSE} & \textbf{Cross Correlation} & \textbf{Power Curve Similarity} \\
        \hline
        12 & 0.320 & 0.464 & 0.625 \\\hline
        24 & 0.228 & 0.782 & 0.750 \\\hline
        72 & 0.209 & 0.756 & 0.802 \\\hline
        \textbf{168} & \textbf{0.189} & \textbf{0.799} & \textbf{0.803} \\
    \hline
    \end{tabular}
    \caption{Metrics comparison of NQF-RNN models trained on varying sequence lengths for the Pyron Wind Farm}
    \label{nqfrnn-comparison}
\end{table}

Based on these results, further analysis was focused on the 168-hour model. The NQF-RNN used to gather the following results was trained using the Adam optimizer over 32 epochs with LSTMs containing 32 hidden nodes, and a two-layer feedforward network with 32 and 16 hidden nodes \cite{adam}. During training, a learning rate of $\num{1e-3}$ and quantile levels [0.01, 0.05, 0.1, 0.5, 0.9, 0.99] were used. The data was processed into batches of size 6 during training, validation, and testing. For Brownian smoothing, a smoothness factor of 0.01 and a drift factor of 0.005 were applied.

\textcolor{black}{Over the span of 10 years of Pyron power generation data, this model obtained a mean RMSE of 0.188 with a standard deviation of 0.011, a mean cross correlation of 0.803 with a standard deviation of 0.026, and a mean power curve similarity of 0.720 with a standard deviation of 0.027, calculated over each year of data.}



\begin{figure}[h!]
    \centering
    \includegraphics[width=1\linewidth]
    {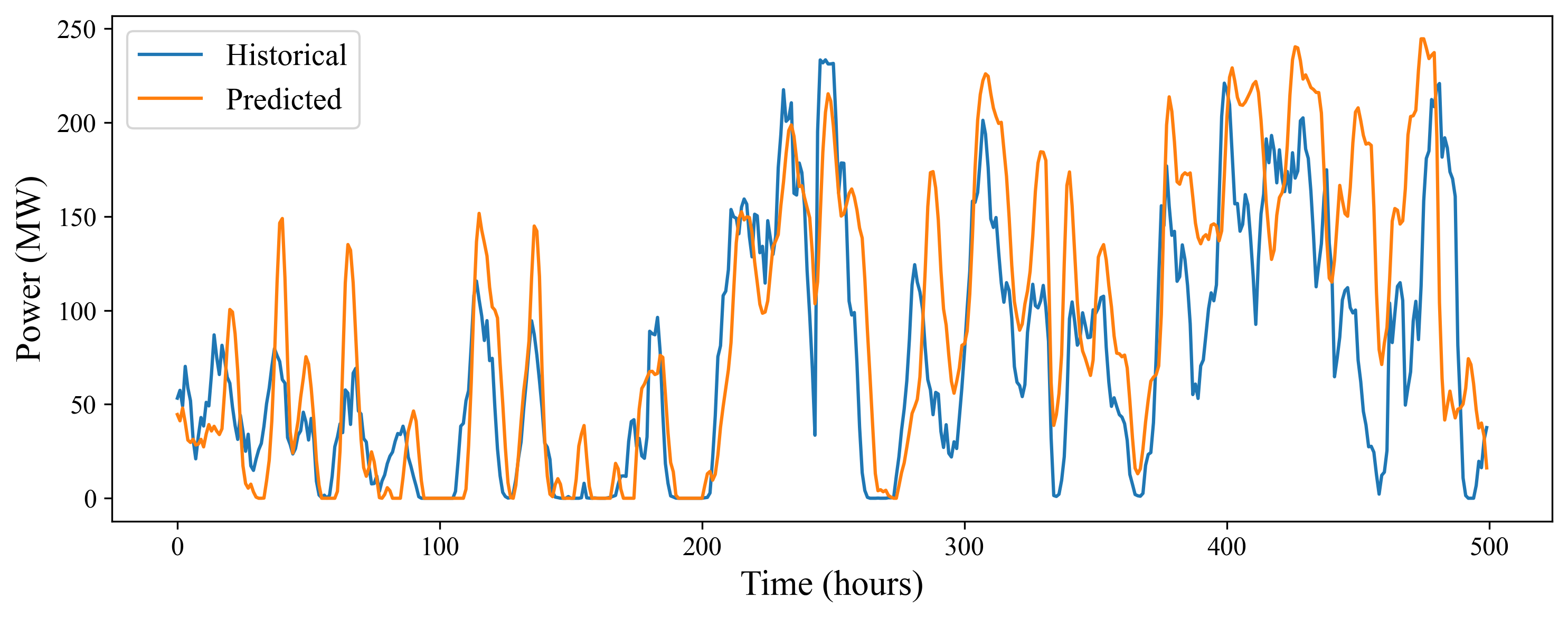}
    \caption{Time series of historical data vs. model predictions}
    \label{fig:nqfrnn-time-series}
\end{figure}

\begin{figure}[h!]
    \centering
    \includegraphics[width=1\linewidth]
    {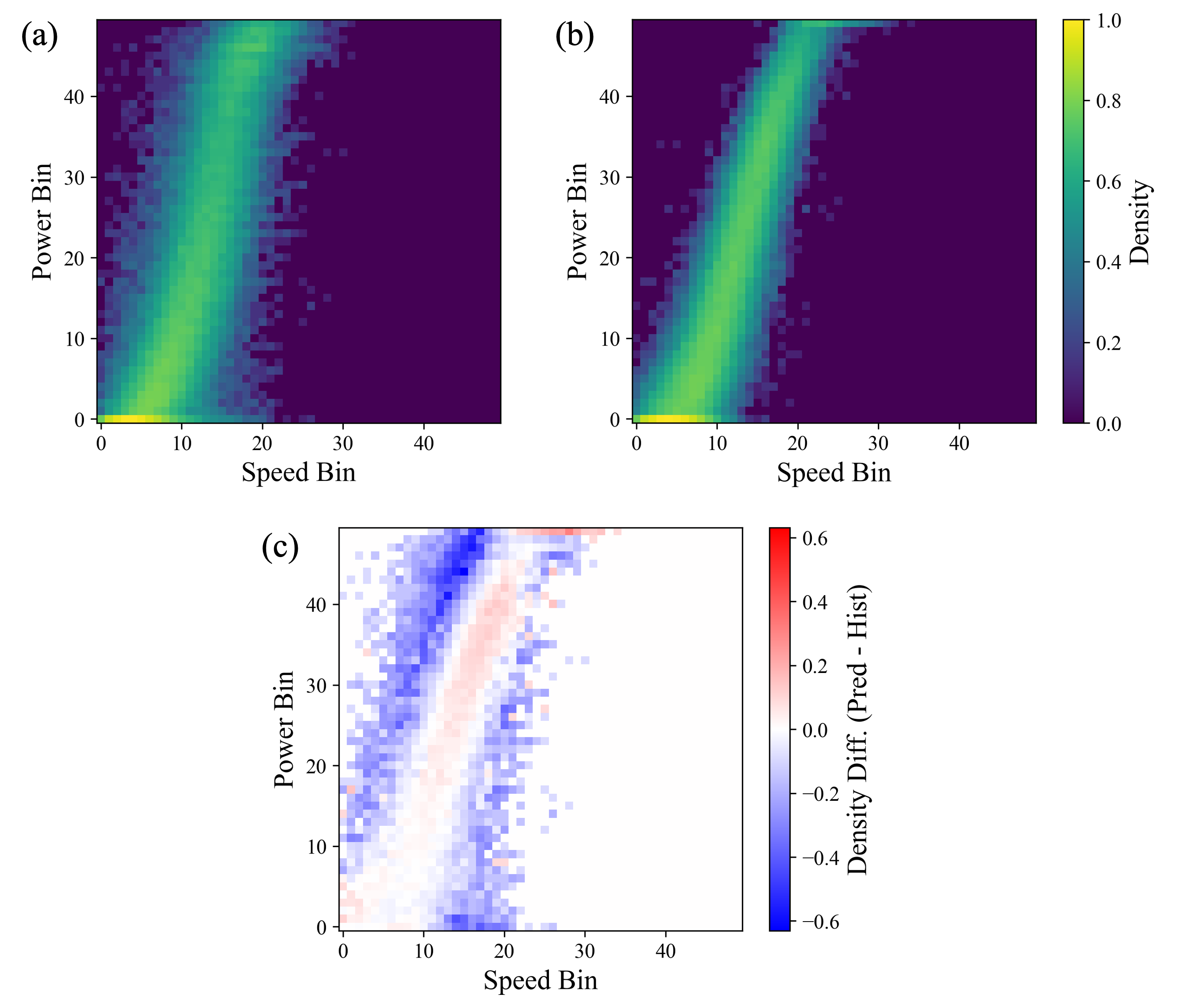}
    \caption{Log-transformed power curve densities. (a) Historical data. (b) Model predictions. (c) Density difference between historical data and model predictions}
    \label{fig:power-densities2}
\end{figure}

Figure \ref{fig:nqfrnn-time-series} shows a section of ERCOT-modeled power generation data from the Pyron wind farm alongside model predictions over wind speeds from the same region in time. The log-transformed power curve densities of the ERCOT-modeled Pyron power generation and predicted power generation were computed against historical wind speed, shown in Figure \ref{fig:power-densities2}. 

Additionally, the 168-hour NQF-RNN model was compared against the WE-Validate baseline models (NWPDB and PLUSWIND) after retraining on the Palouse case study dataset \textcolor{black}{which included 5 months of data}. Table \ref{wevalidate-comparison} summarizes this comparison, with the trained NQF-RNN outperforming both baselines on all evaluation metrics.

\begin{table}[h!]
    \centering
    \begin{tabular}{|c|c|c|c|}
    \hline
        \textbf{Model} & \textbf{RMSE} & \textbf{Cross Correlation} & \textbf{Similarity} \\
        \hline
        NWPDB & 0.230 & 0.808 & 0.577 \\\hline
        PLUSWIND & 0.199 & 0.832 & 0.395 \\\hline
        \textbf{NQF-RNN} & \textbf{0.180} & \textbf{0.848} & \textbf{0.686} \\
    \hline
    \end{tabular}
    \caption{Comparison of NQF-RNN model to the WE-Validate baselines for the Palouse Wind Farm}
    \label{wevalidate-comparison}
\end{table}

The NQF-RNN reduced RMSE by 9.5\% and improved cross correlation by 1.9\% (compared to PLUSWIND), while achieving an 18.9\% improvement in power curve similarity (compared to NWPDB). 

\subsection{Power Dispatch Strategy}
The baseload storage hyperparameter search yielded the results depicted in Table \ref{tab:storage_search}, which prompted the use of compressed air energy storage (CAES) with a rating of 100 MW and a duration of 24 hours for training COVE-NN.

\begin{table}[h!]
    \centering
    \renewcommand{\arraystretch}{1.3}
    \begin{tabular}{|c|c|c|c|}
    \hline 
       {\textbf{Storage Type}} & \textbf{Rating} & \textbf{Duration} & \textbf{Avg. COVE} \\
       & \textbf{(MW)} & \textbf{(hrs)} & \textbf{(\$/kWh/yr)} \\
       \hline
       Lithium-Ion & 100 & 2 & 119.08 \\
       \hline
       Hydropower & 100 & 100 & 134.20 \\
       \hline
       \textbf{CAES} & \textbf{100} & \textbf{24} & \textbf{117.01} \\
       \hline
       Hydrogen & 100 & 24 & 155.10 \\
       \hline
       Gravitational & 100 & 2 & 145.98 \\
       \hline
       Thermal & 100 & 4 & 153.37 \\
       \hline
    \end{tabular}
    \caption{Comparison of Storage Technologies by COVE Metrics. Bolded row indicates the storage configuration with the lowest average COVE across systems.}
    \label{tab:storage_search}
\end{table}

The COVE-NN hyperparameter search identified the best performing hyperparameter values as shown in Table \ref{tab:hp_res}, which were used for the final trained model. The trained COVE-NN used to gather the following results was trained using the Adam optimizer over 32 epochs, with LSTMs with 16 hidden nodes, and a two-layer feedforward network consisting of 128 and 64 hidden nodes \cite{adam}. During training, a learning rate of $\num{1e-4}$ was used, and the training and validation sets were batched into batches of size 8, where each of the 8 samples in a given batch contained $168$ timesteps.

\begin{table}[h!]
    \centering
    \begin{tabular}{|c|c|}
    \hline
        \textbf{Hyperparameter} &  \textbf{Best Performing Value}\\
        \hline
        $\gamma$ & $1.807$\\
        \hline
        $\Gamma$ & $3.288$\\
        \hline
        $\omega$ & $2.702$\\
        \hline
        $\Omega$ & $2.546$\\
        \hline
        $\lambda$ & $1.000$\\
        \hline
        $\Lambda$ & $0.152$\\
        \hline
        $t_a$ & $8$ epochs\\
        \hline
    \end{tabular}
    \caption{COVE-NN Best Performing Hyperparameters}
    \label{tab:hp_res}
\end{table}

\begin{figure}[h!]
    \centering
    \includegraphics[width=0.9\linewidth]{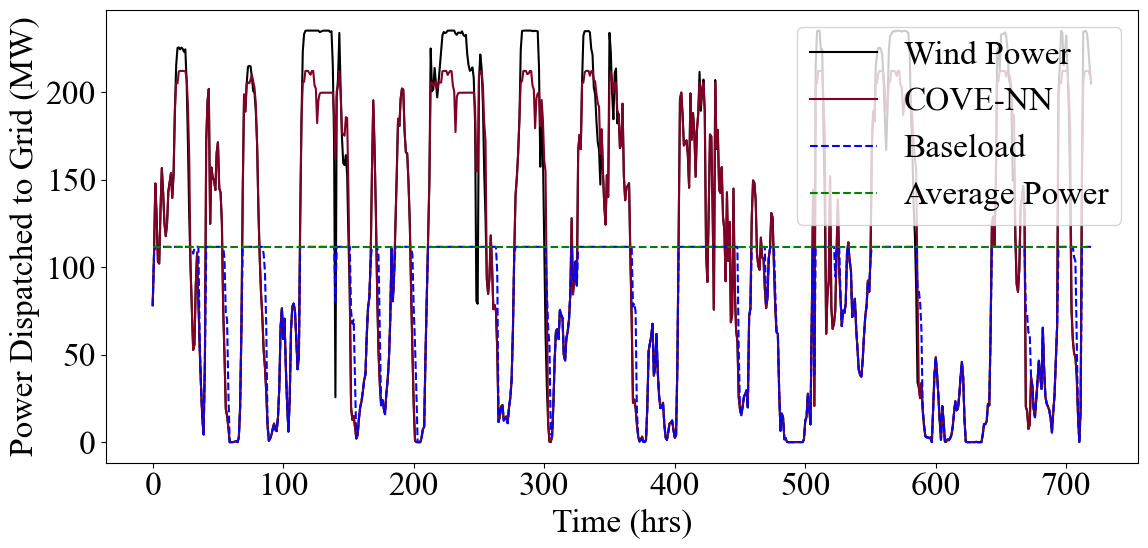}
    \caption{Dispatches from COVE-NN and Baseload over a month of validation power generation data.}
    \label{fig:decisions-cove-nn}
\end{figure}
\begin{figure}[h!]
    \centering
    \includegraphics[width=0.9\linewidth]{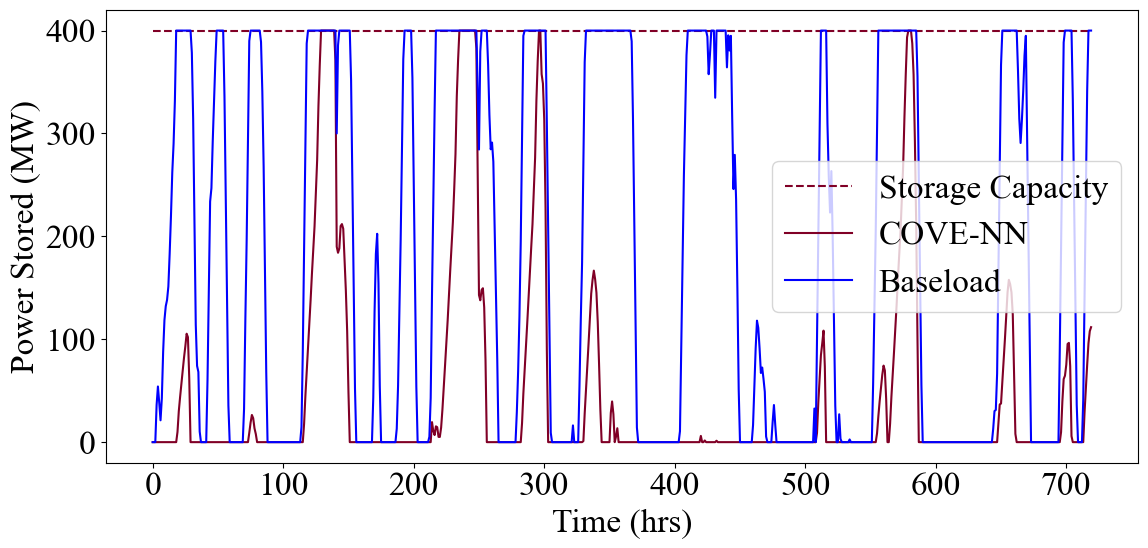}
    \caption{Storage utilization across COVE-NN and Baseload resulting from dispatches depicted in Figure \ref{fig:decisions-cove-nn}.}
    \label{fig:storage-decisions}
\end{figure}

\begin{figure}[h!]
    \centering
    \includegraphics[width=0.8\linewidth]{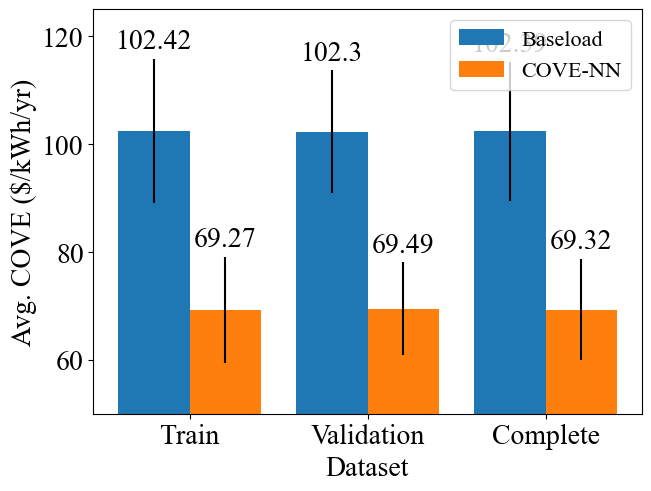}
    \caption{COVE-NN vs. Baseload Performance. \textcolor{black}{These results were computed annually over each of the 43 years in the COVE-NN Pyron case study dataset, and the bars represent the average across all years spanned by each dataset. The error bars represent the standard deviation across all years spanned by each dataset.}}
    \label{fig:cove-compare}
\end{figure}
\begin{table}[h!]
    \centering
    \begin{tabular}{|c|c|c|}
    \hline
        \textbf{Model} & \textbf{Baseload} & \textbf{COVE-NN}\\
        \hline
        Annual Energy Production (MWh) & 512282 & 749656\\
        \hline
        Annual Power Curtailment (MWh) &  246273 & 9647\\
        \hline
        Avg. Annual Storage Utilization (\%) & $40\%$ & $5\%$\\
        \hline
        Value Factor (VF) &  0.992 & 0.993 \\
        \hline
    \end{tabular}
    \caption{Annual Delivery, Curtailment, and Storage Utilization across Models. Note that these results were computed annually across each of the 43 years in the COVE-NN Pyron case study dataset and averaged across all 43 years.}
    \label{tab:cove_stats}
\end{table}
Over the Pyron wind farm case study dataset, COVE-NN finds improvements in COVE over baseload of 32.4\% on the training dataset, 32.1\% on the validation dataset, and 32.3\% on the dataset overall. In Table \ref{tab:cove_stats} and Figures \ref{fig:decisions-cove-nn}, \ref{fig:storage-decisions}, and \ref{fig:cove-compare}, baseload and COVE-NN use the same storage system: CAES with a rating of 100 MW and a duration of 4 hours. Despite the model being trained with a 24 hour storage duration, due to the low storage utilization depicted in Table \ref{tab:cove_stats}, the model duration was decreased as this decreased the capital and operational expenditures associated with the storage system, and did not bottleneck COVE-NN's storage utilization.


\section{Discussion}

\subsection{Power Generation Modeling}
In the power generation modeling case, evaluation of the NQF-RNN model across varying input time series lengths during training revealed that longer temporal contexts increase the model's predictive accuracy. Specifically, the 168-hour model outperformed models trained on shorter sequences, demonstrating the importance of incorporating temporal dependencies in wind power forecasting models.

The application of Brownian smoothing further refined model predictions, resulting in more consistent transitions between predictions across timesteps. Previous iterations of the model without this smoothing procedure produced very noisy predictions, while the historical data was relatively smooth. This smoothing approach is similar to the random walk smoothing used in the Bayesian model BELMM \cite{sarala}, although it has not been applied in previous work on deep learning for wind energy applications.

Comparative analysis of the NQF-RNN model with the baseline models (NWPDB and PLUSWIND) indicated that the NQF-RNN model not only achieved higher accuracy than existing models, but was able to effectively learn data across multiple wind farms with different atmospheric and infrastructure contexts. However, comparing model performance of the Palouse farm with the Pyron farm, there is a notable drop in similarity score from 0.803 to 0.686. This discrepancy may be due to the limited amount of data for the Palouse site, which only had data spanning 6 months compared to the roughly 10 years of historical data available for the Pyron wind farm. Additionally, the similarity metric was not directly optimized during training as it measures distributional similarity, while the NQF-RNN model was designed to optimize pointwise predictions. The higher similarity scores for the Pyron site compared to the Palouse site suggest that the similarity metric becomes more stable with larger datasets, where distributional estimates are less affected by sampling variability.

\textcolor{black}{While the NQF-RNN captures the inherent uncertainty in power generation through its probabilistic output distribution, the current framework assumes that wind speed inputs are known or accurately simulated. In practice, wind speeds are forecasted and are subject to error, introducing additional uncertainty in the input data that can affect model performance. Future extensions could incorporate probabilistic or ensemble wind forecasts to better represent this source of uncertainty.}

\subsection{Power Dispatch Strategy}
Evaluating the trained COVE-NN performance on the Pyron case study dataset suggests that the model is generalizing effectively to unseen data from the Pyron wind farm, as the model is able to achieve a comparable improvement over the baseload dispatch strategy for the training and validation datasets. The average annual COVEs achieved by baseload over the training and validation datasets are very similar, suggesting that the train-validation split resulted in a relatively balanced representation of power generation, market electrical price, and user load behavior across the Pyron case study dataset. That is, outlier years containing massive power outages do not cause significant imbalances across the training and validation datasets.

On average, COVE-NN annually delivers much more power than baseload given the same power generation data as input, but it does so by storing far less energy. COVE-NN annually utilizes on average $5\%$ of its storage capacity, suggesting that it could be more cost effective with a smaller capacity storage system. \textcolor{black}{Such low storage utilization reflects a limitation of COVE-NN, as the model does not take full advantage of the potential impact of on-site energy storage and has decreased robustness against intermittency in the event of exceedingly high energy demands (i.e. during massive outages). In order to better understand COVE-NN storage utilization behavior, case studies should be carried out in future work on additional wind farms with varying degrees of wind penetration.} This lower storage utilization explains the low curtailment incurred by COVE-NN compared to baseload as well. The value factor of both COVE-NN and baseload is close to 1, which suggests that both dispatch strategies tend to release power around a near-constant supply. In the case of baseload, \textcolor{black}{the power dispatched tends towards} the actual average power generation of the Pyron wind farm. In the case of COVE-NN, based on total power delivery and behavior depicted in Figure \ref{fig:decisions-cove-nn}, the \textcolor{black}{power dispatched tends to be greater} than that of baseload, thereby maximizing power delivery while maintaining a high value factor. \textcolor{black}{Thus, COVE-NN learns to dispatch high-value stable power with reduced grid intermittency}. 
\par This high value factor performance is notable in that the model was not directly trained to maximize its value factor. \textcolor{black}{Rather, based on the choice of hyperparameters, early-epoch loss penalties incurred by the model for deviating from baseload directed the in-training model towards peak-shaving, stable dispatch behavior, promoting a high value factor.} That COVE-NN learned to maintain a high value factor suggests the effectiveness of imposing terms in the unsupervised loss function that act similar to Bayesian priors during the initial training. \textcolor{black}{While peaking-mode strategies are more common at present — given that wind penetration typically remains below 10\% and energy-arbitrage operation dominates existing practice — it is equally important to investigate baseload-type dispatch frameworks under future scenarios where wind energy is anticipated to become a mainstream power source exceeding 40\% penetration. In such a high-penetration context, transforming wind power into a stable baseload-like resource becomes both technically and economically relevant, warranting detailed investigation of low-intermittency operational strategies such as those explored by COVE-NN. Future work could extend this work's unsupervised learning methodology around more dynamic peaking-mode dispatch strategies, which are often more effective for maximizing energy value compared to baseload, albeit intermittently.}

\section{Conclusion}
This work develops two novel deep learning frameworks with COVE-NN, a wind power dispatch strategy tailored to individual power systems, and a model for wind power generation modeling from atmospheric data. Both of these frameworks use LSTM-based recurrent neural networks to capture temporal dependencies in their respective input series, which improves power generation modeling performance over baselines in RMSE, cross-correlation, and similarity score over historical data gathered from the Palouse wind farm. For COVE-NN, despite on average utilizing only 5\% of its storage capacity, the model’s power dispatching over 43 years of simulated power generation data from the Pyron wind farm results in a 32.3\% improvement in COVE over baseload. Additionally, COVE-NN was able to achieve a high value factor in its performance over the Pyron case study dataset, which reflects the practical applicability of the model for reducing wind farm intermittency. 

Towards a future integration of COVE-NN with the aforementioned power generation model, this work defines a methodological foundation for individually tailoring power systems' dispatch strategies \textcolor{black}{and power generation modeling}, thereby providing practical tools for strengthening grid connection, \textcolor{black}{reducing grid} intermittency, and improving the economic performance of individual wind farms \textcolor{black}{for a high-penetration renewable future}. These deep learning frameworks could be enhanced by incorporating wind speed forecasting to provide look-ahead features during training, or extended to grid-level dispatch strategies that operate across multiple systems. \textcolor{black}{In particular, a future implementation could integrate the generation and dispatch frameworks into a unified pipeline, enabling synthetic forecasts to drive real-time dispatch decisions.} Such developments would enable the broader applicability of data-driven control methods for renewable energy optimization.

\section*{Acknowledgements}
\addcontentsline{toc}{section}{Acknowledgements}

This work was supported in part by the National Science Foundation (NSF) Research Experiences for Undergraduates (REU) Program under Award No. 2243980. The third author also gratefully acknowledges support from the U.S. Department of Energy (DOE) Visiting Faculty Program (VFP) at Pacific Northwest National Laboratory (PNNL) during the Summer 2025 term, administered by the Oak Ridge Institute for Science and Education (ORISE).

\bibliographystyle{unsrt}
\bibliography{references}

\vspace{12pt}

\end{document}